\documentclass[letterpaper, 10 pt, conference]{ieeeconf}  %

\IEEEoverridecommandlockouts                              %

\usepackage{graphics} %
\usepackage{epsfig} %
\usepackage{amsmath} %
\usepackage{amssymb}  %
\usepackage{subfig}
\usepackage{comment}
\usepackage{algorithm, algorithmic}
\usepackage{url}
\usepackage{color}
\usepackage[noadjust]{cite} %
\usepackage{balance} %
\usepackage{makecell} %
\newcommand{\todo}[1]{}
\renewcommand{\todo}[1]{{\color{red}Todo: {#1}}}

\usepackage[font=small,labelfont=bf]{caption}

\usepackage{xspace}
\newcommand{\methodname}{FormNet\xspace}

\title{\LARGE \bf
Visual Identification of Articulated Object Parts
}

\author{Vicky Zeng$^{1}$, Tabitha Edith Lee$^{1\dagger}$, Jacky Liang$^{1\dagger}$, Oliver Kroemer$^{1}$%
\thanks{$^{1}$Carnegie Mellon University, Pittsburgh, PA 15123, USA. {\tt\small \{vzz, tabithalee, jackyliang, okroemer\}@cmu.edu}}%
\thanks{$^{\dagger}$ Equal Contribution}%
}

\begin{document}

\setlength{\textfloatsep}{0.2cm}
\setlength{\floatsep}{0.2cm}
\maketitle

\begin{abstract}
As autonomous robots interact and navigate around real-world environments such as homes, it is useful to reliably identify and manipulate articulated objects, such as doors and cabinets. 
Many prior works in object articulation identification require manipulation of the object, either by the robot or a human.
While recent works have addressed predicting articulation types from visual observations alone, they often assume prior knowledge of category-level kinematic motion models or sequence of observations where the articulated parts are moving according to their kinematic constraints.
In this work, we propose \methodname, a neural network that identifies the articulation mechanisms between pairs of object parts from a single frame of an RGB-D image and segmentation masks.
The network is trained on 100k synthetic images of 149 articulated objects from 6 categories.
Synthetic images are rendered via a photorealistic simulator with domain randomization.
Our proposed model predicts motion residual flows of object parts, and these flows are used to determine the articulation type and parameters.
The network achieves an articulation type classification accuracy of 82.5\% on novel object instances in trained categories.
Experiments also show how this method enables generalization to novel categories and can be applied to real-world images without fine-tuning.
\end{abstract}

\section{Introduction}
Reliable, autonomous robots have many potential applications as assistants to humans in settings such as homes, businesses, and hospitals~\cite{bohren2011towards, bekey2008status, doelling2014service, dario1996robotics, riek2017healthcare, yang2020combating}.
One prerequisite for these applications is the capability for robots to both recognize and manipulate \textbf{articulated} objects: objects that have moving parts that are kinematically linked with each other, such as doors, windows, drawers, caps, dials, buttons, and switches. 
For example, a robot tasked with fetching medicine must identify and interact with several articulated objects: opening a door to enter a room, searching a cabinet of drawers for a medicine bottle, twisting the bottle cap open, and then delivering the contents. 
Manually specifying articulation constraints for the vast diversity of objects is intractable, so it is important for a robot to autonomously identify these constraints and their parameters.

Interactive Perception (IP) is a well-known approach to this problem~\cite{bohg2017interactive}. 
With IP, the robot interacts with objects in the environment through physical contact, observes the changes, and predicts the underlying kinematic constraints. 
For example, the robot may pull on a handle, and if the handle's trajectory forms a straight line, then the constraint is prismatic (e.g., drawer); if the handle follows an arc, then the constraint is revolute (e.g., door).
Prior IP works often share the limitation of not leveraging the object's \textit{visual} features --- either at all~\cite{sturm2010operating, sturm2011probabilistic, hofer2014extracting, barragan2014interactive} or only as a contextual prior to exploration~\cite{moses2020visual}.
Most articulated objects that humans interact with are designed to visually signify their articulation affordances through pronounced geometry and texture.
An elongated bar with one contact at the end to another surface is probably a revolute handle; a cabinet handle with a connection at each end is likely prismatic.

\begin{figure}[!t]
    \centering
    \subfloat{{\includegraphics[width=.95\linewidth]{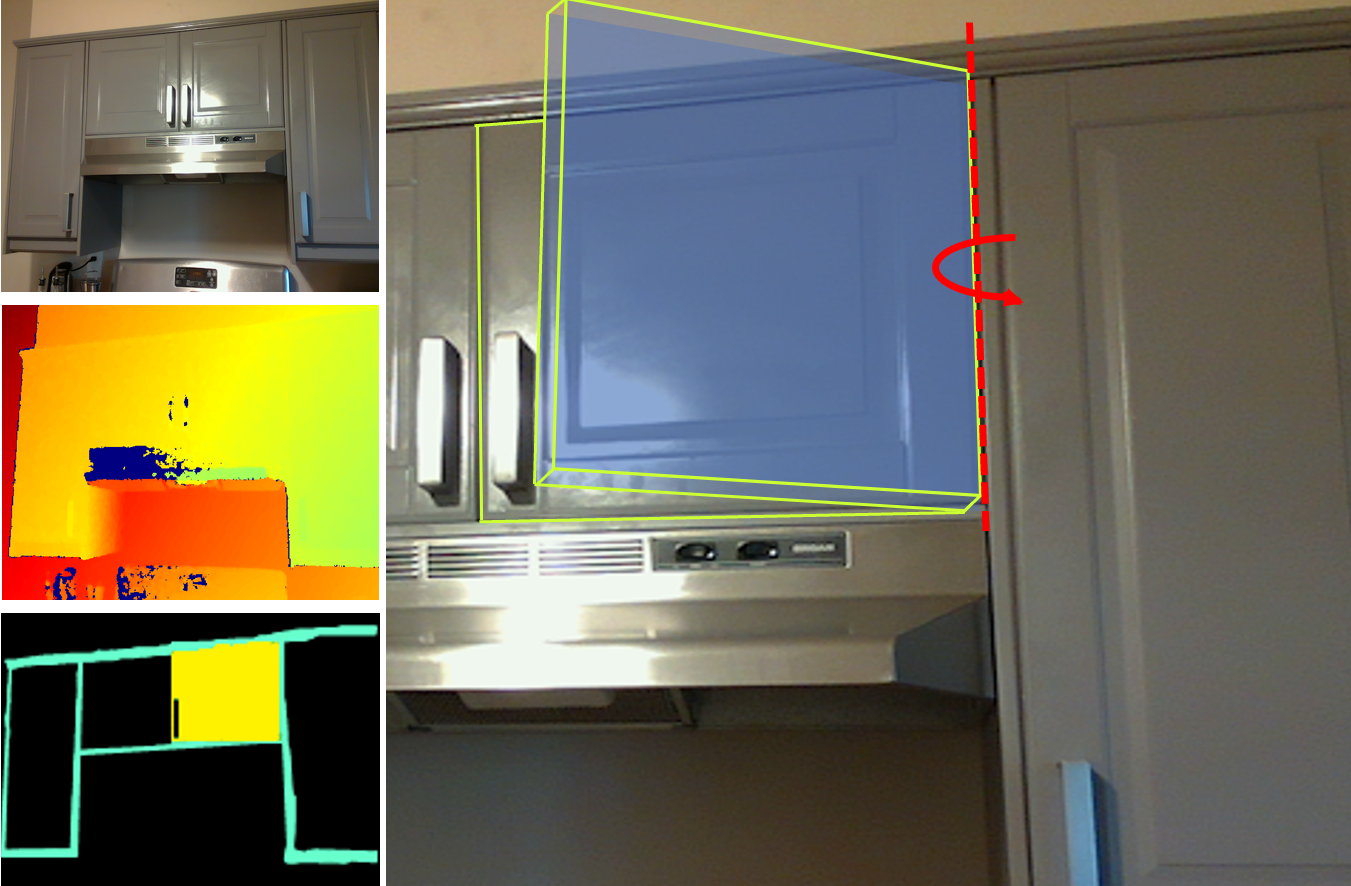}}}%
    \caption{
    Predicting articulation mechanisms using \methodname. 
    Given the inputs on the left (RGB-D image and segmentation masks of a pair of object parts), a neural network predicts the articulation type (revolute, prismatic, fixed, or unconnected) and appropriate articulation parameters (e.g., location and direction of revolute axis).
    The network is trained on synthetic data and infers articulation parameters via predicting motion residual flows.
    }
    \label{fig:teaser}%
\end{figure}

Other approaches consider visual aspects while identifying articulation mechanisms. 
Existing works in this direction track the relative movements of object parts from videos and predict the constraint types~\cite{martin2014online, katz2014interactive, hausman2015active, martin2016integrated, martin2017building, baum2017opening, eppner2018physics}. 
However, these still require the articulated object to be manipulated --- an onerous assumption for robots operating in novel environments. %

The contribution of our work is \methodname\footnote{FormNet stands for \textbf{F}low of \textbf{O}bject \textbf{R}esidual \textbf{M}otion \textbf{Net}work. The name alludes to the \textit{form}, or shape, of object motion under articulation constraints.}: a neural network model that identifies articulation mechanisms between objects with only single-frame vision observations, no interactions, and no pre-specified object category model (Fig.~\ref{fig:teaser}).
We leverage recent developments in high-quality object mesh datasets\footnote{\url{https://sim2realai.github.io/Synthetic-Datasets-of-Objects-Part-I/}} that contain information about both object parts and their relative kinematic constraints.
Color and depth (RGB\nobreakdash-D) images, as well as part segmentation masks, are collected across six object categories found in PartNet-Mobility~\cite{xiang2020sapien}. 
These categories include common household objects like doors, windows, and cabinets.
Training images are rendered in simulation with domain randomization.
To predict articulation type, we represent and encode the displacement of an object part under a given articulation mechanism as motion residual flow.
The network uses convolutions to predict this flow and the parts' ``connectedness'' (i.e., whether two object parts are connected), which we then post-process with RANSAC to form the articulation type prediction.
We evaluate the performance of the trained network with ablation studies across multiple object categories, and we also demonstrate that it can predict articulation types of objects in real-world images without further fine-tuning.
See supplementary materials at \url{https://sites.google.com/view/articulated-objects/home}
\section{Related Work}

Our work relates to two broad classes of vision-based object perception for (1) articulation constraints and (2) pose estimation in manipulation settings.

\textbf{Visual identification of object articulation constraints.}
To identify articulation mechanisms via visual observations, the authors of~\cite{wang2019shape2motion} manually labeled a large dataset with motion parameters, such as the location and axis of revolute and prismatic joints.
Then, they proposed using motion-driven features and losses to train neural networks that jointly solves for motion-driven part segmentation and motion parameters.
Here, motion parameters were encoded as displacement and orientation residuals, corresponding to prismatic and revolute joints.
However, this method assumes access to the \textit{complete} point cloud of an object, not a partial or noisy point cloud that would be found with egocentric depth sensors used in most robotic manipulation applications.

Later works relaxed the assumption of complete point clouds, but they leveraged knowledge of a set of predefined articulated object categories and their kinematic models.
For example, doors would be one category, and cabinets another.
Knowledge of object categories allow these algorithms to fit predefined geometric and kinematic models to the observed visual features, which are often point clouds.
For instance, in~\cite{abbatematteo2019learning} the authors formed Gaussian mixture models over six predefined kinematic models, and they trained a neural network to predict parameters of the mixture model from single depth images.
The parameters include kinematic model parameters for each category, the object's joint configurations, and geometry parameters (e.g., door length).
Training data is generated in simulation, and the model generalizes to novel objects within known categories.

In~\cite{li2020category}, the authors forgo mixture of Gaussians by proposing the articulation-aware normalized coordinate space hierarchy, a
canonical representation for each articulated object category.
Within this representation, object scales, orientations, and articulation parameters are normalized, allowing a neural network to directly regress to coordinates in this space.
The proposed model uses PointNet++ to process point clouds extracted from depth images, and depth image data for training is also generated in simulation.

The authors of ScrewNet~\cite{jain2020screwnet} removed the assumption of known object categories or kinematic models.
Instead, they represent the relative motion of point clouds as a screw motion: rotation of a body around an axis coupled with a translation in the axis.
ScrewNet is a neural network that directly predicts parameters of this screw motion between articulated parts, inferring articulation type without known kinematic models.
However, to make this prediction, the network requires a sequence of depth images, with the articulated object parts moving relative to each other.

Procrustes-Lo-RANSAC (PLR) \cite{huang2012occlusion, huang2014occlusion} similarly predicts articulation types without \textit{a priori} kinematic models.
PLR leverages a geometric vision-based algorithm instead of a neural network, but it requires observations of the object in two distinct articulation configurations.

Like~\cite{jain2020screwnet, huang2012occlusion, huang2014occlusion}, our work does not require prior definitions of category-level kinematic models.
Instead, our proposed method uses \textit{a single image observation} to predict articulation type; no observations of part motion are required.

\textbf{Vision-based object pose estimation for manipulation.}
Our work can be viewed as a form of vision-based pose estimation, wherein we infer the constrained poses that the connected parts of an object would move under the predicted articulation kinematics.
In this view, our work relates to pose estimation in manipulation settings, which includes both objects~\cite{tremblay2018corl:dope, xiang2018posecnn, peng2019pvnet, florence2018dense} and robots~\cite{bohg2014robot, widmaier2016robot, Lee-2020-119511, lu2020robust}.
Of these works, PVNet~\cite{peng2019pvnet} also regresses to a residual (pointing to object keypoints for pose estimation), whereas our motion residual is used to infer the kinematic constraint.
For practical manipulation scenarios, we also note the success of DART~\cite{schmidt2014dart}, a depth-based tracking algorithm for articulated models.
Our learning-based model facilitates single-image estimation of articulation motion that generalizes without needing a model specification for novel objects, as DART does.
Lastly, our system can be integrated into a vision-based manipulation pipeline, such as the one described in~\cite{tremblay2020indirect}. Instead of object pick-and-place, our approach would facilitate object articulated motion, such as opening cabinets and doors.

\section{Method}

In this section, we describe the proposed neural network model (\methodname) for vision-based identification of articulations of object parts, how its training data is generated, and the representations of the model's inputs and outputs.

\begin{figure*}[!t]
  \includegraphics[width=\textwidth]{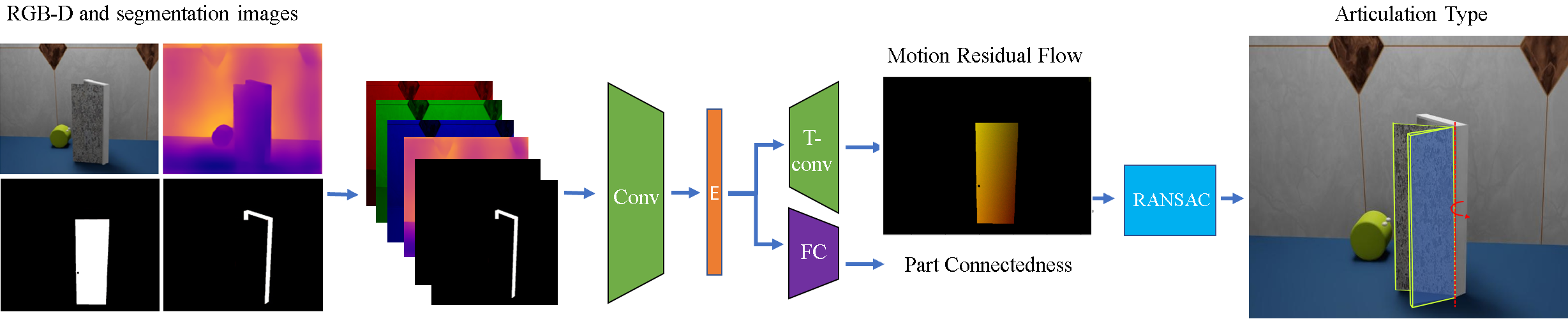}
  \caption{
    Neural network architecture for \methodname.
    It takes as input the RGB, depth, and segmentation images of two queried object parts.
    Conv means convolution layers, E means intermediate embedding, T-conv means transposed convolution layers, and FC means fully-connected layers. 
    The network produces two outputs: motion residual flow and binary part connectedness. 
    If the two queried parts are predicted to be connected, plane fitting via RANSAC is used to post-process the motion residual flows to estimate the articulation mechanism's type and parameters.
    }
  \label{fig:model}
\end{figure*}

\subsection{Overview}

Our approach identifies kinematic constraints between pairs of object parts from a stationary visual observation alone.
We construct a neural network model that takes as input a single-view RGB\nobreakdash-D image and segmentation masks of two distinct object parts.
We focused on designing and training this network, and assume the part segmentation masks are provided from a pre-existing algorithm (e.g.,~\cite{wang2019voxsegnet}).
The neural network provides two outputs for the segmented parts: (1) a parts connectedness classification and (2) the motion residual flow, which is the displacement of the second part relative to the first part if the second part moves under their kinematic constraint.
Two parts are connected if they are parts of the same articulated object and are neighbors of one another in the object's kinematic chain.
The joint for two connected parts is classified as revolute, prismatic, or fixed, depending on the predicted motion residual flow.
Furthermore, we apply RANSAC on the predicted motion residual flow to extract articulation parameters for revolute (rotation axis) and prismatic (direction of movement) joints.

By formulating our network to infer the articulation type between a pair of object parts, our approach works with images containing an arbitrary number of articulated objects.
Therefore, predefined categorical models or shared coordinate spaces of articulated objects are not needed; the network does not need to reason about multiple articulation mechanisms belonging to specific object categories.
In addition, this formulation also allows our model to generalize to object categories unseen during training.
The entire kinematic chain of an articulated object in an image can be recovered by querying the network with all pairs of object parts.

The proposed model is trained via a large dataset of synthetically rendered images of articulated objects.
Domain randomization and augmentations are applied to the training images, facilitating a network that is robust and invariant to changes in viewpoint, lighting, textures, occlusions, and object joint configurations.
Models of articulated objects came from PartNet-Mobility, from which we filtered and cleaned models to form our training set.

\subsection{Dataset of Articulated Objects}

\begin{table}[!t]
\begin{center}
\begin{tabular}{|c|c|c|c|c|c|c|}
\hline
 & \textbf{\textit{Door}} & \textbf{\textit{Window}} & \textbf{\textit{Faucet}} & \textbf{\textit{Dishw.}} & \textbf{\textit{Fridge}} & \textbf{\textit{Cab.}}\\
\hline
\textbf{Objects} & 30 & 28 & 23 & 30 & 20 & 18\\
\hline
\textbf{Parts} & 92 & 112 & 115 & 92 & 98 & 109\\
\hline
\textbf{Type} & R & P & R & R & R & P,R \\
\hline
\end{tabular}
\caption{
    Dataset Statistics. 
    \textbf{Objects} shows the number of object models.
    \textbf{Parts} shows the total number of distinct object parts. 
    \textbf{Type} lists the type of articulations that exist in that category, where R = revolute and P = prismatic.
    \textbf{Dishw.} stands for Dishwasher, and \textbf{Cab.} stands for Cabinet.
}
\label{tab:data_stat}
\end{center}
\end{table}

We considered several public datasets of objects with object part information for making our training data, including RBO~\cite{martin2019rbo}, ShapeNet~\cite{chang2015shapenet}, and PartNet~\cite{mo2019partnet}. 
To train a generalizable articulation identification model, an object dataset is needed that contains a wide variety of object categories, a large number of diverse objects within each category, and labels of articulation types between connected object parts.
The RBO dataset is a collection of $14$ objects with $358$ RGB-D interaction video sequences. 
While it provides articulation and part segmentation, the relatively  small size makes it inadequate as training data. 
ShapeNet consists of over $3$ million 3D CAD models, of which $220$K are classified to $3135$ categories.
PartNet contains roughly $26$K models across $24$ object categories with good part segmentation. 
While ShapeNet and PartNet are sufficiently large, both datasets lack articulation information.

In recent works, Shape2Motion~\cite{chang2015shapenet} and PartNet-Mobility~\cite{xiang2020sapien} have augmented ShapeNet and PartNet to include articulation information. 
Shape2Motion is large with over $2.4$K objects.
However, it is not amenable to simulation and rendering; it lacks joint limit information and object textures.
PartNet-Mobility does not face such limitations and has our desired properties: the dataset has over $2.4$K objects across $46$ categories, object textures, and articulation information with joint limits.
As such, we use PartNet-Mobility to generate the training data for our articulation prediction model (Table~\ref{tab:data_stat}).

We processed meshes from PartNet-Mobility to choose (1) categories with strong visual signals in kinematic constraints and (2) characteristic subsets of objects from each chosen category. 
The six categories chosen were doors, windows, faucets, dishwashers, refrigerators and cabinets. 
In each category, we choose a  representative subset of objects that maintains intra-category variability. 
In total, $149$ objects were selected. 
We cleaned the selected mesh models by scaling meshes to realistic sizes and standardizing the orientation of their coordinate frames.
The former makes rendered images more realistic (i.e., faucets are typically smaller than refrigerators).
The latter ensures that objects of the same categories appear in similar poses when loaded for rendering.
Lastly, we removed object parts and articulation connections that were too detailed.
For example, the interior racks of dishwasher models were ignored in our dataset.

\subsection{Dataset of Scene Images with Articulated Objects}

The network is trained using synthetic data with domain randomization and image augmentation. 
Data was generated using NVIDIA Isaac Sim\footnote{\url{https://developer.nvidia.com/isaac-sim}}, a GPU-accelerated robotics simulator that supports photorealistic rendering.
Articulated objects were loaded into a clean virtual scene and several randomizations were applied, including camera pose, scene lighting, object pose, size, texture, and distractor objects.

To make the trained articulation prediction model generalizable across a wide variety of scenes, we perform domain randomizations to render the synthetic dataset.
The camera is positioned randomly within the front upper hemisphere of the object, a region where a robot is likely to be to perform manipulation.
Joint configurations of articulated objects are randomized within their joint limits. 
Object sizes are also randomly scaled between $0.5$ and $2.0$ during generation. %
In addition, we randomize object textures and scene lighting.
These randomizations accommodate for intra-category variation. While household items have diverse appearances (i.e., the width, color, and length of doors might differ), these particular differences do not affect the underlying articulation mechanisms.
Scaling and changing the visual properties of the objects in the training data makes the model invariant to such details.
Lastly, distractor objects consisting of common objects and household items are included to produce natural occlusions of articulated object parts.

In total, approximately $100$K image scenes were rendered, each including an RGB-D image of resolution $640\times 480$, an object part segmentation image, and the articulation information of objects in the scene.

In addition to domain randomization, we apply standard image augmentation techniques during training, including geometric transformations such as random rotations, flips, and crops~\cite{info11020125}.
For RGB images, we perform random visual transformations such as contrast and brightness.
We also add realistic noises to depth and segmentation masks to make the trained model more robust. 
For depth images, we apply additive correlated Gaussian noise and multiplicative gamma noise to simulate realistic depth sensor noise~\cite{mahler2017dex}.
To add realistic noise to the boundaries of the binary segmentation masks, we apply salt and pepper noise followed by a morphological closing operation.

Each data sample during training is generated with a pair of object parts in a rendered image.
We first pick a pair of object parts from the segmentation image.
The network input then consists of the RGB-D image of the entire scene and two binary segmentation masks of the chosen pair of object parts.
The output consists of a binary part-connectedness label of the corresponding pair as well as the motion residual flow of the second part relative to the first part.
The motion residual flow is a $W\times H\times 3$ image, where each pixel is only non-zero if it occupies a pixel belonging to the second object part.
See Figure~\ref{fig:dataset} for some examples.
In these non-zero pixels, the values of each pixel correspond to where the corresponding point on the object part in 3D space would be if the object part is moved by a fixed magnitude following its kinematic constraint with the other part.
The direction of the motion is expressed in the camera frame.
For fixed joints, the motion residual is $0$.
For revolute joints, the movement magnitude is $30^\circ$.
For prismatic joints, the movement magnitude is $0.3M$, where $M$ is the maximum joint movement distance provided by PartNet-Mobility.  
The exact magnitudes of these movement offsets are not important, since the network is not tasked with learning the range of motion of articulated objects, just their articulation types.

\subsection{Network Architecture}

The neural network for our approach is an hourglass encoder-decoder architecture similar to the network used for DREAM~\cite{Lee-2020-119511}. 
As shown in Fig.~\ref{fig:model}, the network takes as input a stacked image observation of size $640\times 480\times 6$, with $4$ RGB-D channels and $2$ part-segmentation mask channels.
The network predicts the motion residual flow as a $640\times480\times3$ image and binary part-connectedness label.

The image encoder consists of the convolutional layers of VGG19 pretrained on ImageNet~\cite{SimonyanZ14a}. 
The decoder (upsampling) module has four 2D transpose convolutional layers (stride = $2$, padding = $1$, output padding = $1$), and each layer is followed by a normal $3\times 3$ convolutional layer and ReLU activation layer. 
The first output head is the part connectedness, consisting of 3 fully connected layers.
The second output head for the motion residual flow is composed of 3 convolutional layers ($3\times3$, stride = $1$, padding = $1$) with ReLU activations with $64$, $32$, and $3$ channels, respectively.
There is no activation layer after the final convolutional layer. 

The network is trained with a Cross Entropy loss on part connectedness and a Mean Square Error loss on the motion residual flow.
Let $y^c_n\in \{0, 1\}$ and $y^f_n \in \mathbb{R}^{640\times 480\times 3}$ respectively denote the binary part-connectedness and motion residual flow of the $n$th training sample, and $\hat{y}^c_n, \hat{y}^f_n$ be their estimated counterparts produced by the neural network.
The weighted loss function on the two outputs is defined as:
\begin{align}
    \mathbf{SE}(y^f_n,\hat{y}^f_n) &= \|y^f_n - \hat{y}^f_n\|^2 \\
    \mathbf{CE}(y^c_n, \hat{y}^c_n) &= -y^c_n\log(\hat{y}^c_n) + (1-y^c_n)\log(1-\hat{y}^c_n) \\
    \mathcal{L} &= \frac{1}{N}\sum_{n=1}^{N} w_{se} y^c_n \mathbf{SE}(y^f_n,\hat{y}^f_n) + w_{ce} \mathbf{CE}(y^c_n, \hat{y}^c_n)  
\end{align}
Note that we only propagate the motion residual squared error loss when the parts are connected.
The weights $w_{se} = 0.6$ and $w_{ce} = 0.4$ were chosen after hyperparameter search.

\subsection{Articulation Prediction from Motion Residual Flows}

\begin{figure}[!t]
    \centering
    \subfloat{{\includegraphics[width=.48\linewidth]{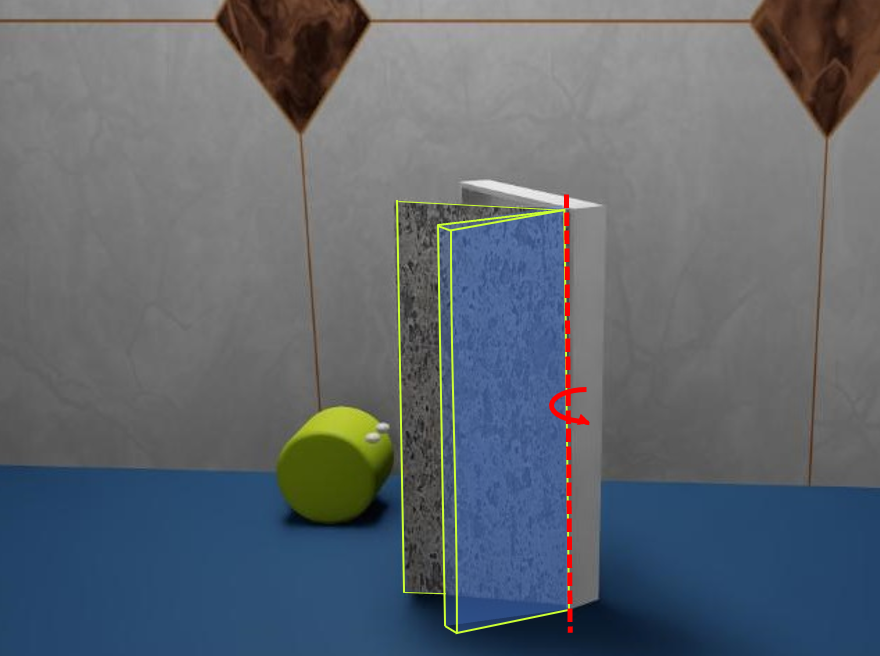} }}
    \subfloat{{\includegraphics[width=.48\linewidth]{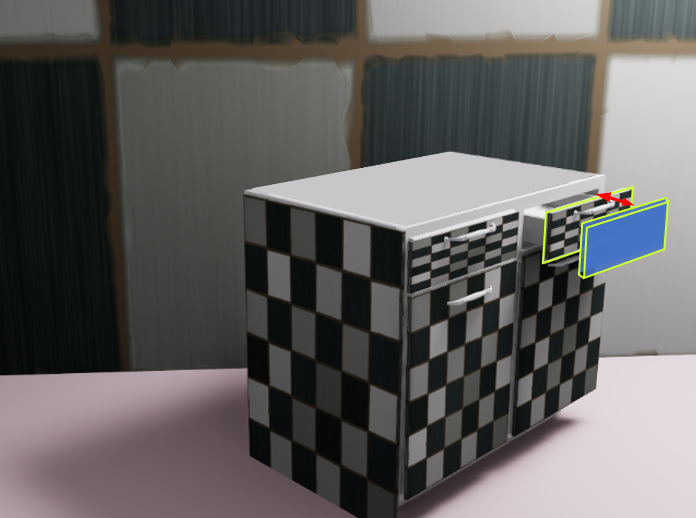} }}
    \caption{
        Visualizations of estimated pre-motion and post-motion planes. 
        In both images, pre-motion planes are outlined in bright green on the original object part, while the post-motion planes are visualized with a blue infill.
        In the left image, the red annotations denote the estimated axis of the predicted revolute joint.
        In the right image, the red arrow denote the direction of the prismatic joint.
    }
    \label{fig:planes}
\end{figure}

If the network predicts that the two queried object parts are connected, then we process the predicted motion residual flow to robustly estimate the part articulation type and parameters.
Algorithm pseudocode can be found in the supplementary materials.
First, a plane is fitted on the point cloud of the object part in the input observation.
We refer to this as the pre-motion plane.
Next, we fit a second plane on the point cloud of the object part, where each point is translated by the predicted motion residual flow.
We refer to this as the post-motion plane.
Refer to Fig.~\ref{fig:planes} for examples of revolute and prismatic planes. 

For fitting both planes, RANdom Sample Consensus (RANSAC)~\cite{fischler1981random} is used to obtain robust estimations given depth and segmentation noise that exists in the network inputs as well as estimation errors in the network outputs.

The articulation type and parameters can be inferred through comparison of the position and orientation of the pre-motion and post-motion planes.
If the planes are sufficiently close together (i.e., the predicted motion residuals are all close to $0$), then the predicted articulation type is fixed.
Otherwise, if the pre-motion and post-motion planes are sufficiently parallel, then the articulation type is prismatic.
In this case, the direction of the prismatic kinematic constraint is the direction of the average motion residual flow.
Lastly, if the motion residuals are not all close to $0$ and the planes are not parallel, then the predicted articulation type is revolute.
In this case, the axis and location of the revolute joint is the line where the pre-motion and post-motion planes intersect.
Extracting articulation parameters from motion residual planes in this manner allows the network to learn a single output representation that works for fixed, revolute, and prismatic joints.

\section{Experiments}

\begin{figure*}%
    \centering
    \includegraphics[width=\linewidth]{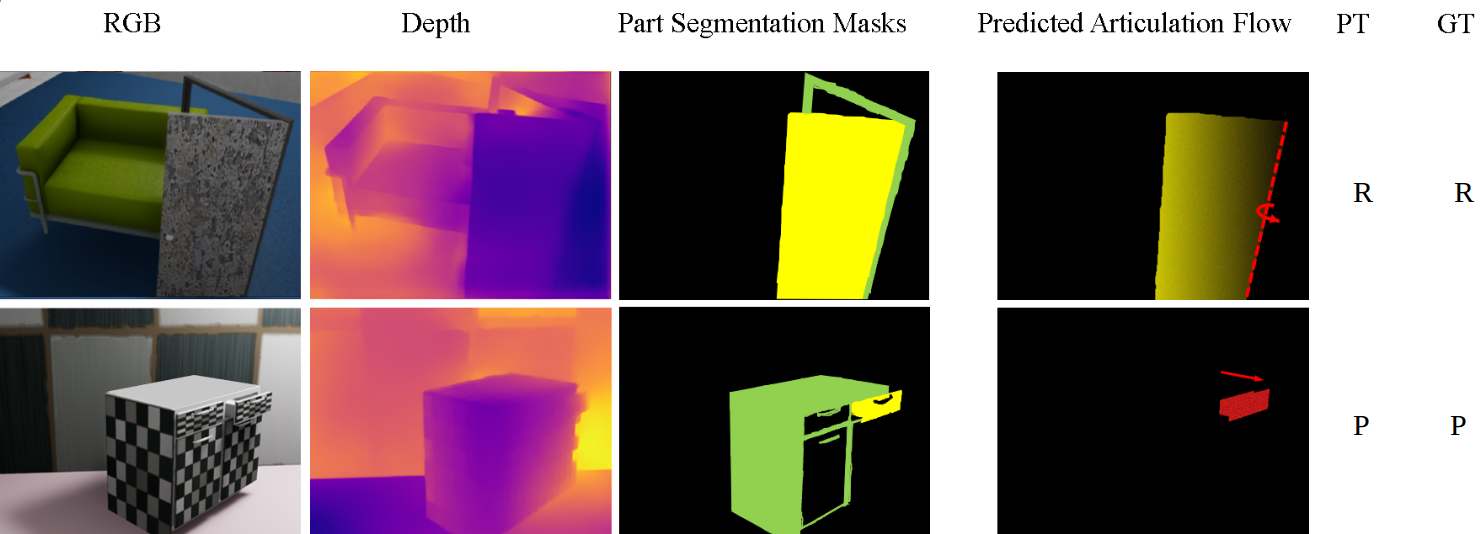}
    \includegraphics[width=\linewidth]{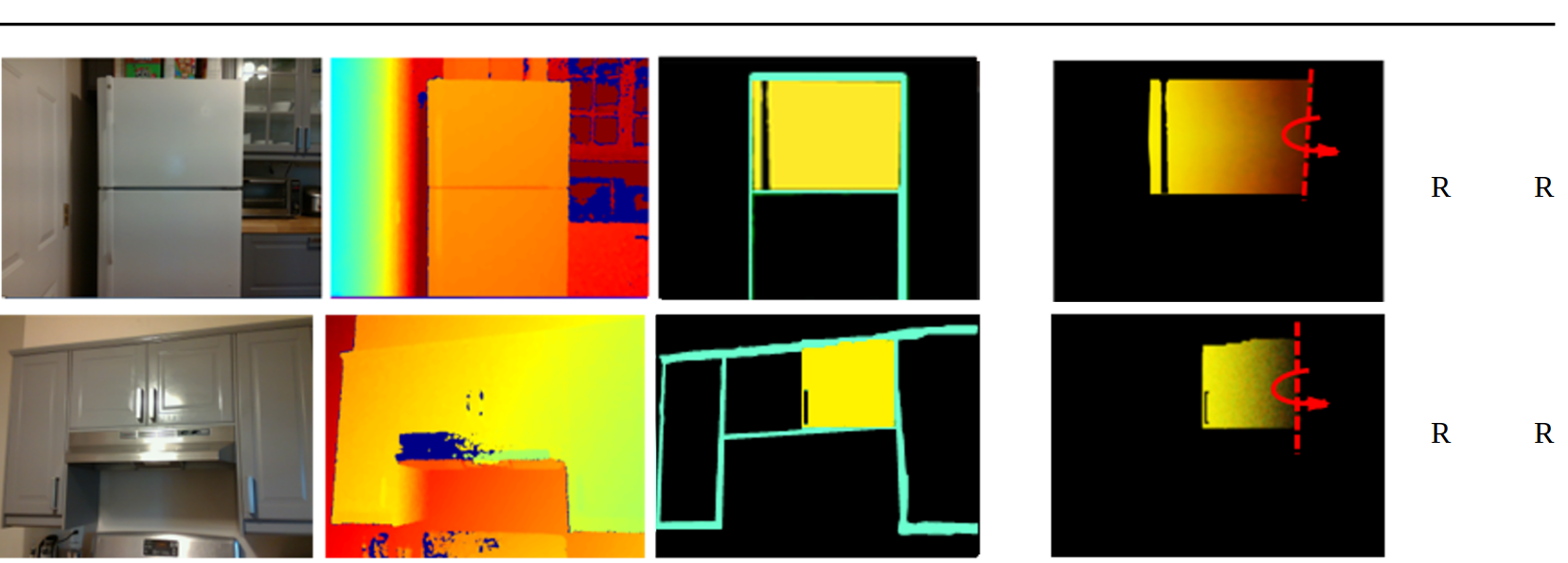}
    \caption{
    Visualization of network inputs and outputs for the Door and Cabinet categories on synthetic data (top two rows) and real-world images (bottom two rows). 
    Additional categories are visualized in supplementary materials.
    PT means predicted articulation type, and GT ground truth type. 
    Articulation is either revolute (R), prismatic (P), fixed (F), or unconnected (U).
    The network takes in a stack observation of RGB image, depth image, and two part segmentation masks. 
    The green segmentation mask is the anchor object part and the yellow segmentation mask is the candidate object part. 
    The direction of the motion residual flow is visualized by color gradients, where a prismatic articulation is a single solid color, while a revolute articulation is a gradient towards the axis of rotation.
    }
    \label{fig:dataset}
\end{figure*}

We evaluate our network on synthetic images and show successful transfer to real-world data. 
Specifically, we report (1) test accuracy achieved with our network with synthetic images when trained on all object categories, (2) generalization to categories unseen during training in a leave-one-out fashion, and (3) generalization to real-world images. 
See Fig.~\ref{fig:dataset} for representative qualitative results.
An ablation study is further conducted to train the network on each single category and test against all others, to analyze knowledge transfer between categories.

\subsection{Network Training}
The neural network was implemented with PyTorch and optimized via the Adam optimizers with a learning rate of $1.2\times10^{-4}$ and momentum of $0.9$.
These were tuned via hyperparameter search.
The training set consisted of $70$k images, and the remaining $30$k were used in the test set.
The network took $30$ epochs to train, taking 32 hours on an NVIDIA Tesla V100 GPU.

\subsection{Network Accuracy (All Object Categories)}
To evaluate the accuracy of the classified articulation types, we separate the predictions into $4$ classes: prismatic, revolute, fixed, and unconnected. We refer to this as combined accuracy (\textbf{CA}).
Two additional metrics are evaluated: accuracy over part connectedness (\textbf{PC}) and accuracy over connected articulation type (\textbf{AT}).
The former is the binary classification accuracy of whether or not two parts are connected.
The latter is only the articulation type accuracy when the network predicts a true positive part connectedness.

\begin{figure}%
    \centering  
    \includegraphics[width=0.49\linewidth]{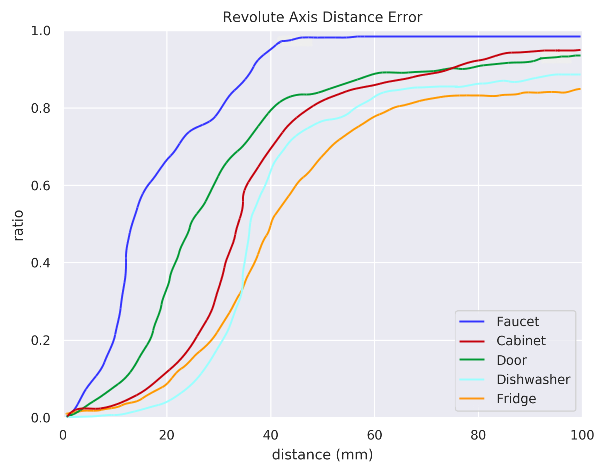}
    \includegraphics[width=0.49\linewidth]{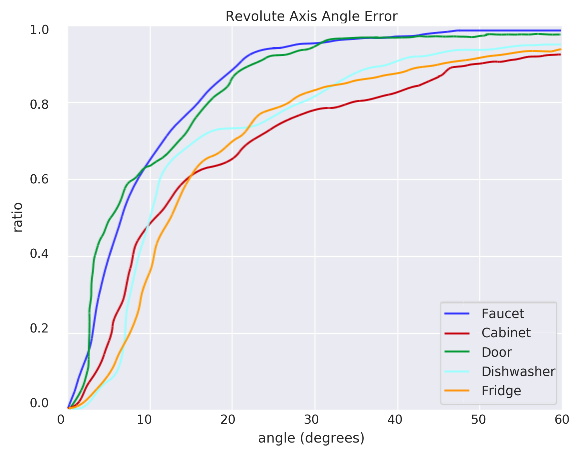}
    \includegraphics[width=0.6\linewidth]{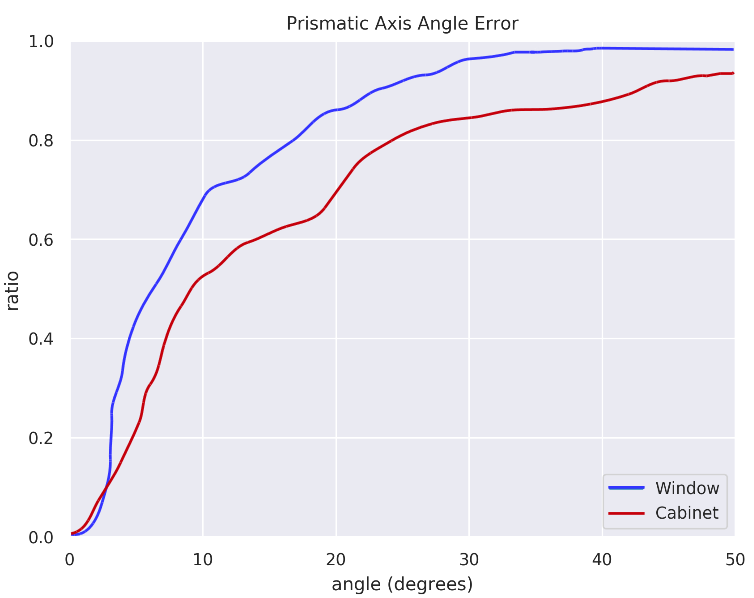}
    \caption{
    Predicted articulation parameter accuracy for revolute axes (top) and prismatic axes (bottom) on test set for all object categories.
    The revolute axis distance error is the average distance between the points on the ground truth axis to their projections on the predicted axis.
    The revolute and prismatic angle errors are the angle between the ground truth and the predicted axes.
    All plots show the percentage of data points that have error below a given threshold.
    }
    \label{fig:rev_add_dist}
\end{figure}

Table~\ref{tab:perf} shows the test classification metrics by object category when the network is trained on all object categories.
Similar test accuracies are achieved across categories. 
Fig.~\ref{fig:rev_add_dist} shows the accuracy of the predicted articulation parameters (location and direction of revolute axes, and direction of prismatic axes).

\begin{table}%
\begin{center}
\begin{tabular}{|c|c|c|c|c|c|c|c|}
\cline{2-8}
\multicolumn{1}{c|}{} & \multicolumn{7}{c|}{\textbf{Object Category}} \\
\hline
 \textbf{Metric} & \textbf{\textit{Door}} & \textbf{\textit{Wind.}} & \textbf{\textit{Fauc.}} & \textbf{\textit{Dish.}} & \textbf{\textit{Frid.}} & \textbf{\textit{Cab.}} & \textbf{\textit{Avg.}}\\
\hline
\textbf{AT} & $85.6$ & $75.1$ & $84.4$ & $72.1$ & $76.2$ & $67.7$ & $76.4$ \\
\hline
\textbf{PC} & $97.8$ & $91.2$ & $94.6$ & $95.7$ & $95.4$ & $92.5$ & $94.1$ \\
\hline
\textbf{CA} & $88.7$ & $79.6$ & $87.5$ & $78.6$ & $84.2$ & $77.5$ & $82.5$\\
\hline
\textbf{B-CA} & $81.5$ & $58.7$ & $74.3$ & $60.1$ & $62.7$ & $71.2$ & $68.0$\\
\hline
\end{tabular}
\caption{
    Accuracy by Category on Test Set.
    \textbf{B-CA} refers to a classification-only baseline method where the output head of the network is performing articulation and connectedness classification directly instead of regressing to motion residual flows.
    All numbers are shown in percentage.
}
\label{tab:perf}
\end{center}
\end{table}

\subsection{Generalization to Novel Object Categories}
We also evaluate how well the proposed method generalizes to novel object categories unseen during training.
See Table~\ref{tab:perf_unseen} for results.
We use a leave-one-out scheme for this evaluation. 
Specifically, we train six additional models, with the same hyperparameters, such that each model is trained on all categories except a left out category.
Each model is then evaluated on the category it was not trained on.

\begin{table}%
\begin{center}
\begin{tabular}{|c|c|c|c|c|c|c|}
\cline{2-7}
\multicolumn{1}{c|}{} & \multicolumn{6}{c|}{\textbf{Object Category}} \\
\hline
 \textbf{Metric} & \textbf{\textit{Door}} & \textbf{\textit{Window}} & \textbf{\textit{Faucet}} & \textbf{\textit{Dishw.}} & \textbf{\textit{Fridge}} & \textbf{\textit{Cabinet}}\\
\hline
\textbf{AT} & $78.8$ & $26.5$ & $66.2$ & $76.3$ & $60.3$ & $33.5$\\
\hline
\textbf{PC} & $94.6$ & $96.7$ & $85.7$ & $95.$5 & $93.6$ & $89.2$\\
\hline
\textbf{CA} & $82.1$ & $47.2$ & $73.0$ & $81.48$ & $73.8$ & $58.6$\\
\hline
\end{tabular}
\caption{
    Performance on Novel Object Categories.
    }
\label{tab:perf_unseen}
\end{center}
\end{table}
We observe worst performance on windows and cabinets, mediocre performance on faucets and refrigerators, and best performance on doors and dishwashers.
The difference between the best and worst performing category is significant, differing by over $50$\% (windows at $26.5$\% vs doors at $78.8$\%).
The prediction for windows was worse than chance.

We investigate the abnormally low AT performance of windows in Table~\ref{tab:perf_unseen}, which mainly consists of sliding prismatic joints.
High PC across all categories shows that the model identifies connected parts of windows, but it is predicting the wrong articulation types. 
Analyzing the outputs indicate that most wrong predictions are revolute. 
We hypothesize two reasons for this: (1) lack of representation of prismatic objects in the training data and (2) misleading revolute objects that share visual similarities with the prismatic windows. 
Eliminating windows from the training data leaves five categories, of which only one comprises of prismatic-dominant joints. 

\begin{figure}%
    \centering
    \subfloat{{\includegraphics[width=.6\linewidth]{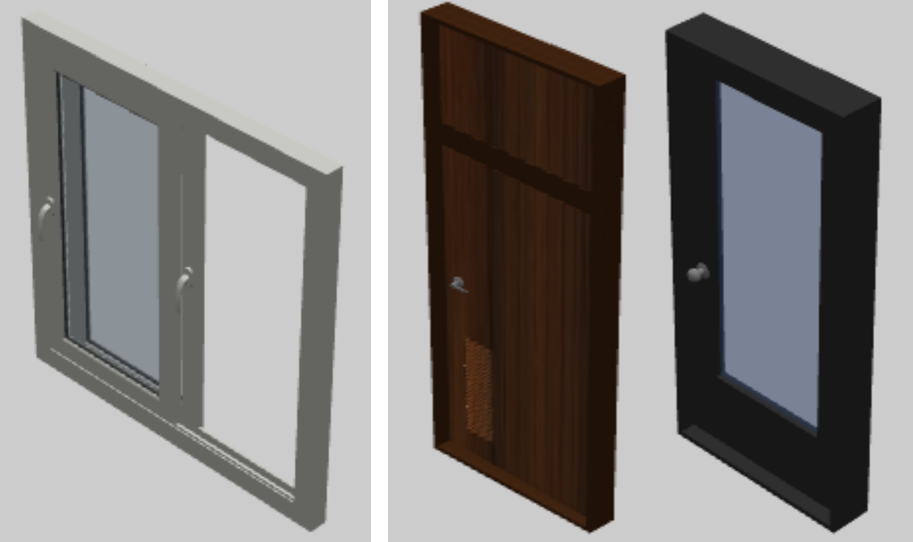} }}
    \caption{
        Misleading examples of a window (left) and two doors (right). 
        The shapes of the door panel and window frames, and handles for the window and first door (brown) are similar.
    }
    \label{fig:door_window}
\end{figure}

We also observe similar visual features across windows, doors, and refrigerators, with the latter two as revolute objects. 
Specifically, they share similar frames, boards and handles, as observed in Fig~\ref{fig:door_window}. 

\begin{figure}%
    \centering
    \subfloat{{\includegraphics[width=.99\linewidth]{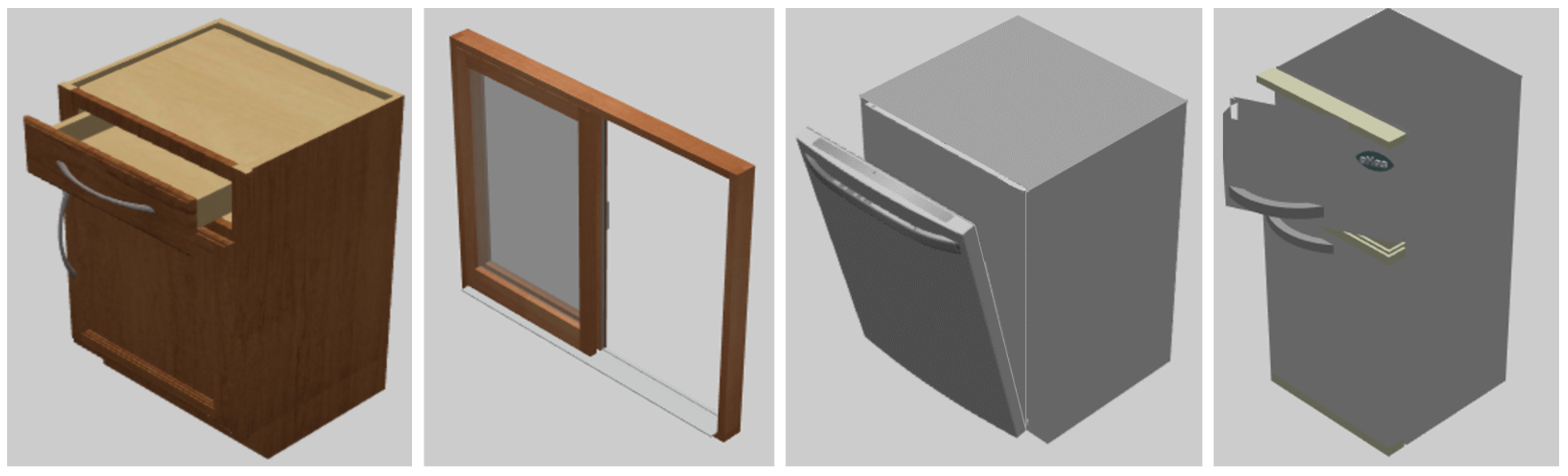} }}
    \caption{
        Misleading examples of a prismatic cabinet (left) vs other objects (prismatic window, revolute dishwasher, refrigerator). 
        The geometry of the cabinet drawer is different from that of the window, but they both have prismatic joints.
        By contrast, the handle of the drawer has a similar shape to those of the dishwasher and the refrigerator, even though the latter have revolute joints.
    }
    \label{fig:cab_comp}
\end{figure}

High visual similarity between windows and revolute objects in the training set may explain the high prediction rate of revolute for windows.  
The importance of visual similarity is further reflected in cabinets, the other low-performing category. 
Although cabinets consist of both prismatic and revolute joints, 73\% of the errors occur on the prismatic slots. 
There was only one training category that was prismatic-dominant (windows), and their object parts showed conflicting visual features with the cabinets as seen in Fig~\ref{fig:cab_comp}.

\subsection{Generalization from Training on One Category}
We formed hypotheses to explain the low transfer for certain unseen categories in the previous section. 
A lack of or misleading similarity in visual features of the object parts resulted in misclassifications. 
This could be significant in our understanding of category generalization.
What features and representation do we need in our training categories for good generalization? 
To answer the question and test our hypotheses, we perform an additional ablation study – training networks on just a single category and testing them across all the other five categories. 
See Table~\ref{tab:perf_one_cat} for results. 

\begin{table} %
\begin{center}
\begin{tabular}{|c|c|c|c|c|c|c|}
\hline
\textbf{Train}&\multicolumn{6}{|c|}{\textbf{Test Category}} \\
\cline{2-7} 
 \textbf{Category} & \textbf{\textit{Door}} & \textbf{\textit{Window}} & \textbf{\textit{Faucet}} & \textbf{\textit{Dishw.}} & \textbf{\textit{Fridge}} & \textbf{\textit{Cab.}}\\
\hline
\textbf{\textit{Door}}& 98.7 & 2.5 & 48.7 & 53.7 & 49.6 & 37.3\\
\hline
\textbf{\textit{Window}} & 3.2 & 93.2 & 8.9 & 7.8 & 5.6 & 38.2\\
\hline
\textbf{\textit{Faucet}} & 33.7 & 8.7 & 88.7 & 4.1 & 4.3 & 17.2\\
\hline
\textbf{\textit{Dishw.}} & 34.2 & 12.8 & 3.8 & 98.5 & 52.1 & 34.6\\
\hline
\textbf{\textit{Fridge}} & 56.2 & 15.5 & 23.2 & 55.2 & 87.5 & 38.7\\
\hline
\textbf{\textit{Cab.}} & 49.6 & 32.5 & 13.5 & 3.4 & 38.7 & 83.2\\
\hline
\end{tabular}
\caption{
    Performance of Training on One Category.
    Overall articulation accuracies when trained on one category (each row) and tested on other categories (each column).
}
\label{tab:perf_one_cat}
\end{center}
\end{table}

Learning is limited between articulation mechanisms. 
Different articulation mechanisms (prismatic, revolute) cannot learn well from each other. 
When trained on a category with only one articulation mechanism, predicting objects with different articulation is low.
For instance, a model trained on windows (prismatic) performs below $10$\% for all but cabinets (the only other category that contains prismatic joints).
We also observe relatively high, consistent performance when testing cabinets, regardless of the trained category.
Limited learning between articulation mechanisms explain this better performance because only cabinets contain both articulation mechanisms. 
Therefore, cabinets have partial representation in each trained category. 
Consequently, cabinets also fail to achieve at least $40$\% for any trained category, due to the presence of both articulation mechanisms. 
Having both articulation mechanisms imply that one of the articulations would be unseen in the train category. 

The results also support our hypothesis that misleading similarity in visual features create misclassifications. 
The lowest transfer occurs between doors and windows. 
While both performs well on its own category, correct prediction of the other category falls in the $2-3$\% range.
Comparing two sample objects from the two categories show similar handles, boards and frames in Fig.~\ref{fig:door_window}.
However, their ground truth articulation types are different, which explains the low performance. 
Dishwashers and refrigerators have the highest transfer, and comparing two sample objects show similar handles and physical structure in Fig.~\ref{fig:cab_comp}.

\subsection{Real-world Experiments}
Despite only being trained using synthetic data, our network also bridges the reality gap when deployed in the real world.
To assess how well our model transfers to real-world data, we took $18$ RGB-D images of various household items in our homes (comprising of refrigerators, doors, faucets, and cabinets).
Part segmentation masks were generated with semi-automatic DEXTR segmentation~\cite{Man+18}.
Results showed successful transfers, where the model predicts the correct articulation types for $12$ of the $18$ images.
Example visualizations are shown in the bottom rows in Fig~\ref{fig:dataset}.

\section{Conclusion} %
We present \methodname, a deep learning approach that predicts articulation mechanisms of object parts from a single image observation without physical interactions and pre-specified categorical kinematic models.
Training data is generated with a photorealistic simulator with $6$ object categories and $149$ objects.
Domain randomization over camera poses, lighting, object sizes, textures, and occlusions make the trained network robust to these variations.
Experiment results show that our approach generalizes to novel object categories in simulation and can be applied to real-world images without fine-tuning.

\section*{Acknowledgment}

This work is supported by the NSF Graduate Research Fellowship Program Grant No. DGE 1745016, NSF Grants No. IIS-1956163  and  CMMI-1925130, the Office of Naval Research Grant No. N00014-18-1-2775, the NVIDIA NVAIL Program, and the CMU Summer Undergraduate Research Fellowship.

\clearpage
\balance
\bibliographystyle{IEEEtran}
\bibliography{citations}

 \begin{appendices}
\normalsize

\section{}

\restylefloat{algorithm}
\begin{algorithm}[H]
\caption{Compute Articulation Type and Parameters from Predicted Motion Residual Flow}
\label{alg:articulation}
\begin{algorithmic}
    \renewcommand{\algorithmicrequire}{\textbf{Input:}}
    \renewcommand{\algorithmicensure}{\textbf{Output:}}
    \REQUIRE Depth image $I_D \in \mathbb{R}^{W\times H}$, binary part segmentation mask $I_S \in \mathbb{Z}_2^{W\times H}$, motion residual flow $I_F \in\mathbb{R}^{W\times H\times3}$, small thresholds $\epsilon_0, \epsilon_1$.
    \ENSURE Articulation Type (AT) $\in$ \{FIXED, PRISM, REV\} and articulation parameters if PRISM or REV.
    \IF {$\|I_f\|_2 < \epsilon_0$ }
        \RETURN FIXED
    \ENDIF
    \STATE Set original point cloud $P \leftarrow$ DEPROJECT$(I_D[I_S])$
    \STATE Set estimated displaced point cloud $P' \leftarrow P + I_F[I_S]$
    \STATE Set pre-motion plane and normals $\pi, \hat{n} \leftarrow$ RANSAC$(P)$
    \STATE Set post-motion plane $\pi', \hat{n}' \leftarrow$ RANSAC$(P')$
    \IF {$\hat{n}^\top\hat{n}' > 1 - \epsilon_1$}
        \STATE Find mean flow $d \leftarrow \frac{1}{\sum_{w, h} I_S[w, h]}\sum_{w, h} I_F[w, h]$
        \STATE Normalize into direction $\hat{d} \leftarrow \frac{d}{\|d\|_2}$
        \RETURN PRISM, $\hat{d}$
    \ELSE
        \STATE Find intersecting line $\mathbf{l} \leftarrow$ INTERSECT$(\pi, \pi')$
        \RETURN REV, $\mathbf{l}$
    \ENDIF
\end{algorithmic}
\end{algorithm}

\section{}

\begin{table}[H]
\begin{center}
\begin{tabular}{|c|c|c|c|}
\hline
\textit{\textbf{Dataset}} & \textit{\textbf{Categories}} & \textit{\textbf{Objects}} & \textit{\textbf{Info}} \\
\hline
\textbf{RBO}~\cite{martin2019rbo} & 14 & 14 & Y \\
\hline
\textbf{ShapeNet}~\cite{chang2015shapenet} & 3315 & 220K & N \\
\hline
\textbf{PartNet}~\cite{mo2019partnet} & 24 & 26.6K & N \\
\hline
\textbf{Shape2Motion}~\cite{wang2019shape2motion} & 45 & 2.4K & Y \\
\hline
\textbf{PartNet-Mobility}~\cite{xiang2020sapien} & 46  & 2.3K & Y \\
\hline
\end{tabular}
\caption{
    This table summarizes the different public datasets of meshes on their number of object categories, number of object models, and whether it contains articulation information between object parts.
    Column \textbf{info} represents articulation information (y/n).
}
\label{tab:datasets}
\end{center}
\end{table}

\section{}

\begin{figure*}[h]
    \includegraphics[width=.968\linewidth]{ims/row_11.png}
    \includegraphics[width=\linewidth]{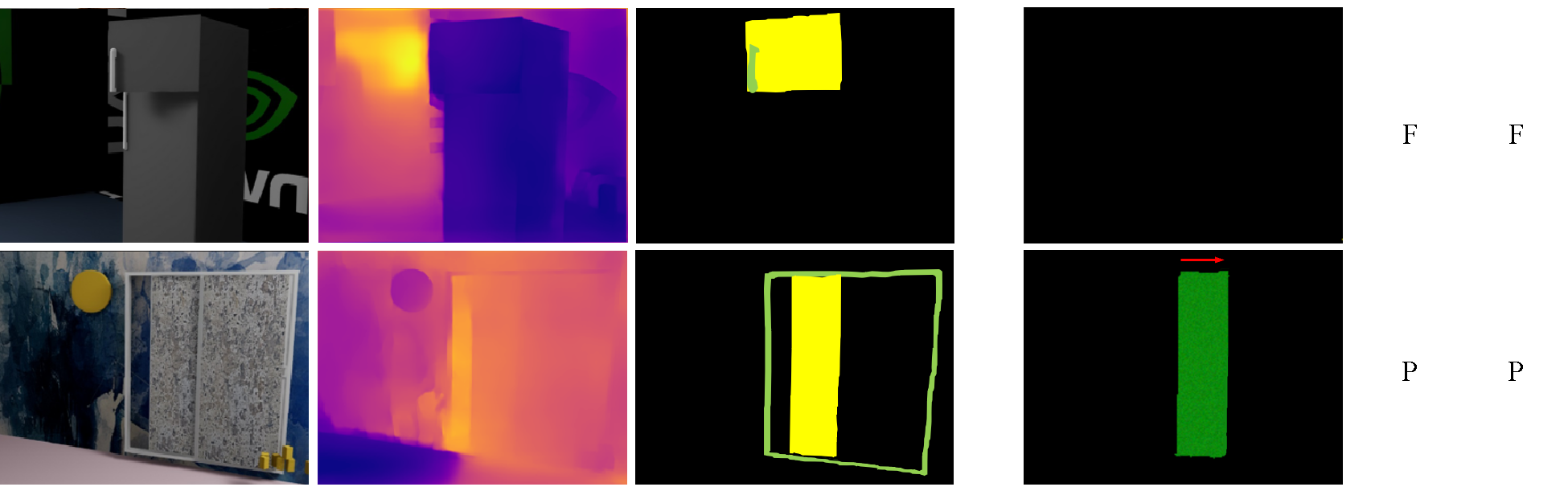}
    \includegraphics[width=\linewidth]{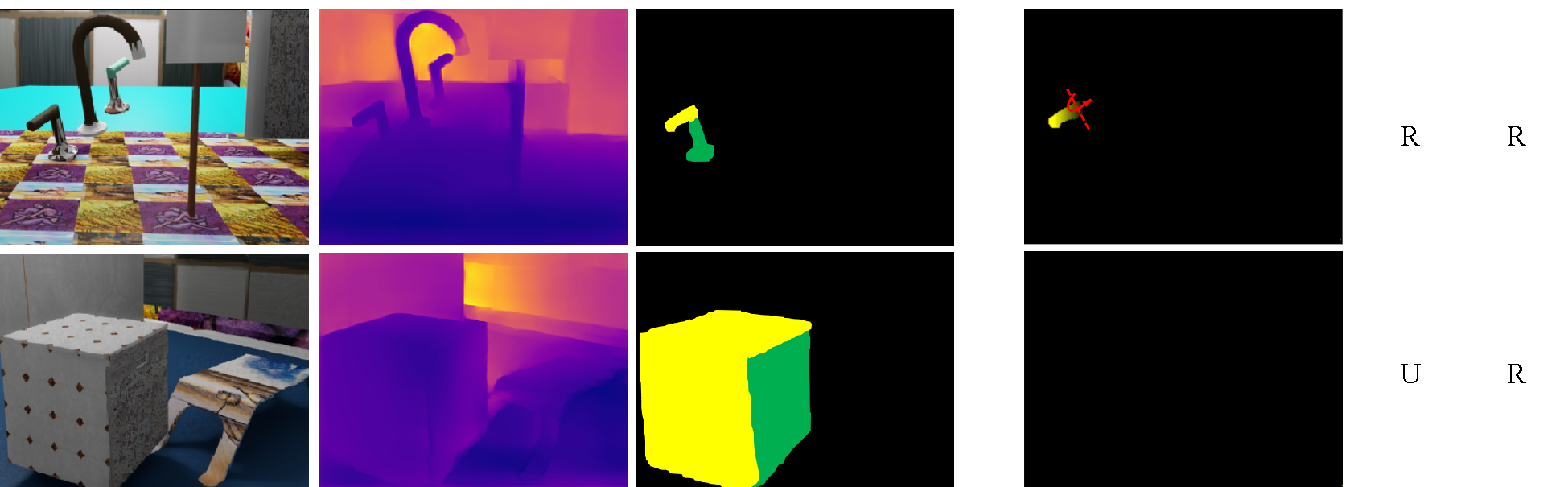}
    \caption{
    Extension of Figure~\ref{fig:dataset} with all object categories.
    }
    \label{fig:dataset_appendix}
\end{figure*}

See Figure~\ref{fig:dataset_appendix} for an extension of Figure~\ref{fig:dataset} with all the object categories.

\end{appendices}

\end{document}